\documentclass[letterpaper]{article} 
\usepackage{aaai23}  
\usepackage{times}  
\usepackage{helvet}  
\usepackage{courier}  
\usepackage[hyphens]{url}  
\usepackage{graphicx} 
\usepackage{natbib}  
\usepackage{caption} 
\usepackage{algorithm}
\usepackage{algorithmic}
\usepackage{newfloat}
\usepackage{listings}
\DeclareCaptionStyle{ruled}{labelfont=normalfont,labelsep=colon,strut=off} 
\floatstyle{ruled}
\newfloat{listing}{tb}{lst}{}
\floatname{listing}{Listing}
\title{Trash to Treasure: \\ Using text-to-image models to inform the design of physical artefacts}
\author{
    Amy Smith\textsuperscript{\rm 1\equalcontrib},
    Hope Schroeder\textsuperscript{\rm 2\equalcontrib},
    Ziv Epstein\textsuperscript{\rm 2},
    Michael Cook\textsuperscript{\rm 3},
    Simon Colton\textsuperscript{\rm 1,4},
    Andrew Lippman\textsuperscript{\rm 2}
}
\affiliations{
    \textsuperscript{\rm 1}School of Electronic Engineering and Computer Science, Queen Mary University London\\
    \textsuperscript{\rm 2}MIT Media Lab\\
    \textsuperscript{\rm 2}Department of Informatics, King's College London\\
    \textsuperscript{\rm 4}SensiLab, Faculty of Information Technology, Monash University\\
    
    amy.smith@qmul.ac.uk, hopes@mit.edu, zive@mit.edu, mike.cook@kcl.ac.uk, s.colton@qmul.ac.uk, lip@media.mit.edu
}

\begin{document}

\maketitle

\begin{abstract}
Text-to-image generative models have recently exploded in popularity and accessibility. Yet so far, use of these models in creative tasks that bridge the 2D digital world and the creation of physical artefacts has been understudied. 
We conduct a pilot study to investigate if and how text-to-image models can be used to assist in upstream tasks within the creative process, such as ideation and visualization, prior to a sculpture-making activity. Thirty participants selected sculpture-making materials and generated three images using the Stable Diffusion text-to-image generator, each with text prompts of their choice, with the aim of informing and then creating a physical sculpture. The majority of participants (23/30) reported that the generated images informed their sculptures, and 28/30 reported interest in using text-to-image models to help them in a creative task in the future. We identify several prompt engineering strategies and find that a participant's prompting strategy relates to their stage in the creative process. We discuss how our findings can inform support for users at different stages of the design process and for using text-to-image models for physical artefact design.
\end{abstract}

\section{Background and Motivations}

Text-to-image deep learning models are generative AI techniques that synthesize images from text inputs. Implementations of this technology such as Midjourney, DALLE-2, and Stable Diffusion \cite{rombach2022high} have exploded in popularity in the last year. 
The company Stability AI has released the models, weights, and code for Stable Diffusion, allowing it to be used publicly, and Midjourney's Discord interface has reached over 1 million users. Increased access to this technology has accelerated the development of open-source tools and resources for creativity and design. As a result, we have seen artists using AI-generated imagery as part of their visual design process. Two examples include a \textit{Cosmopolitan} magazine cover designed in tandem with DALLE-2, \cite{cosmo}, and an artist using Midjourney to win an art competition \cite{wapo}. This democratization has raised questions regarding how such models can be used in a wide variety of tasks, such as idea visualization \cite{EPSTEIN2020101515, epstein2022happy, rafner, smith2022artbhot}.

These tools offer new possibilities for navigating the creative process. Design research suggests a need for flexible pathways for creative computational assistance \emph{early} on in the design process \cite{pip}. Many AI-empowered co-creative tools focus on support for the later stages of design, when participants exploit, refine, or implement ideas, but are less involved in the earliest stages of co-creation, when participants are in the ``explore" phase \cite{angel}.

In contrast to using the generative model for directly creating digital media content, a growing body of work explores the use of generative AI for upstream tasks in the creative process, such as ideation and visualization, that have a range of downstream tangible outcomes. These can include tattoos \cite{tattoos}, fashion \cite{fashion}, community values \cite{epstein2022happy}, and visualizations of the future \cite{rafner}. A key challenge for using generative AI in upstream creative tasks, such as ideation and idea visualization, is the fact that the affordances of physical materials could differ from those of generated imagery, and users may be frustrated bridging that distance to bring their idea into reality. Work by \citet{dang2022prompt} shows that lack of support in the trial and error process of prompt engineering (tactics for refining prompts to synthesize desired outputs) can be frustrating for users, motivating further investigation into users' prompting strategies. Connecting a user's needs, based on their prompting strategy, to their design stage could help provide individualized support for a user's design goals.

\section{Study Design}

We conduct a pilot study to examine the impact of introducing AI-generated images into the early stages of a design task in a physical medium. The experiment was advertised as a community activity at a research university to ``turn trash into treasure'' by making artistic sculptures out of discarded materials. Once participants had joined the activity, they were were instructed by the facilitator to choose 3-5 pieces of material from a box with a range of objects like test tubes, pipe cleaners, foam pieces, and wires.

Once participants had chosen their sculpture materials, we explained that they would give the facilitator three text prompts which would generate three images. We asked them to consider how these images could inform their sculpture design.
We then asked participants for their first prompting phrase, which the facilitator used as input for Stable Diffusion. The facilitator then asked the participant the following whilst the image was being generated: 1) ``Why did you choose that prompt?" and 2) ``What are you expecting to see?" Once the facilitator had written down the responses, the generated image was revealed to the participant. The facilitator repeated this process of prompting, reflection, and image reveal up to another two times. After the visualization stage, the participant was given 3 minutes to build a sculpture using their existing materials, as well as adhesives like tape and hot glue. Once the sculpture had been completed, the facilitator asked 1) ``Was your sculpture informed by your generated images?"  and 2) ``Would you use a text-to-image model like Stable Diffusion for a creative task again?" 

 After the activity, we set out to measure each participant's level of conceptual exploration through their prompting journey. The concept of semantic distance is a popular one in the evaluation of the creative processes \cite{KENETT201911} and is one we believe extends to exploration through semantic space in the early creative process. To operationalize this, we transformed user prompts into sentence embeddings,\footnote{We used the sentence transformer: https://huggingface.co/sentence-transformers/all-MiniLM-L6-v2{\textit{`all-MiniLM-L6-v2'}} from HuggingFace due to its performance capturing semantic meaning.} then measured the cosine distance between a user's first, second, and third prompts, taking the average distance over these to characterize a user's level of conceptual exploration from prompt to prompt. We also qualitatively analyzed the notes from each participant interview as well as the sculptures each produced. 

\section{Results}
\subsection{Generated images informed final designs}
\begin{figure}[h]
  \caption{Visual elements inform sculpture design of a building.}
  \centering
  \includegraphics[width=85mm,scale=0.4]{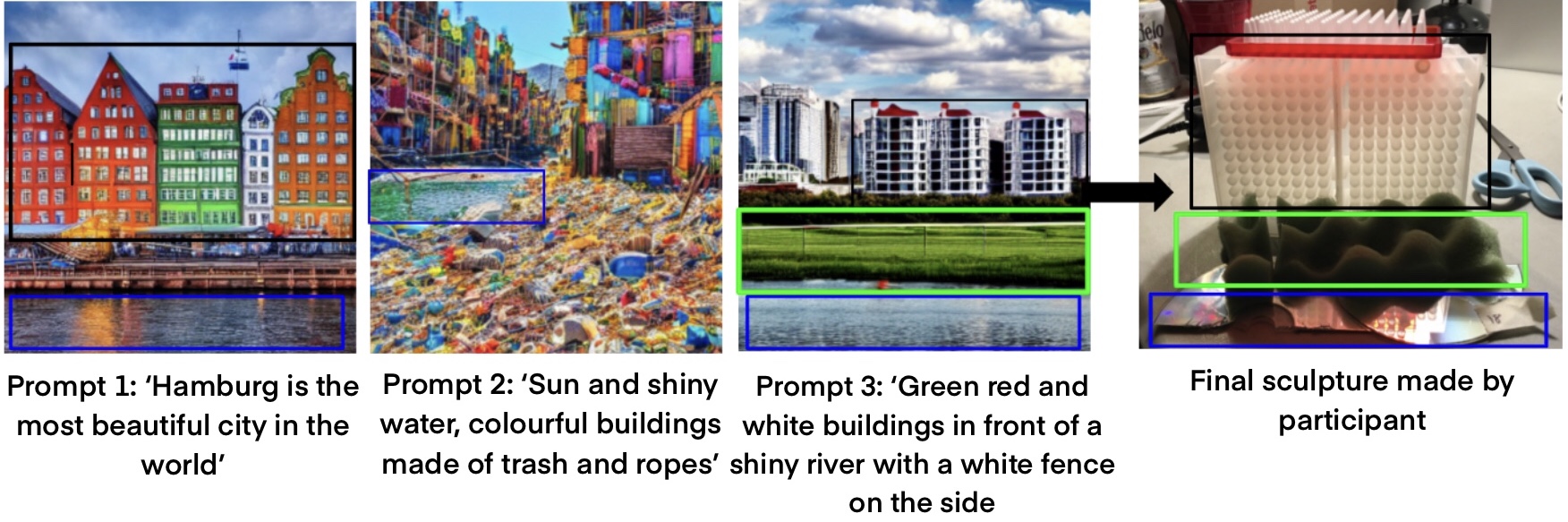}
  \label{fig:building}
  \end{figure}

\begin{figure}[h]
  \caption{Visual elements inform sculpture design of a crab.}
  \centering
  \includegraphics[width=85mm,scale=0.4]{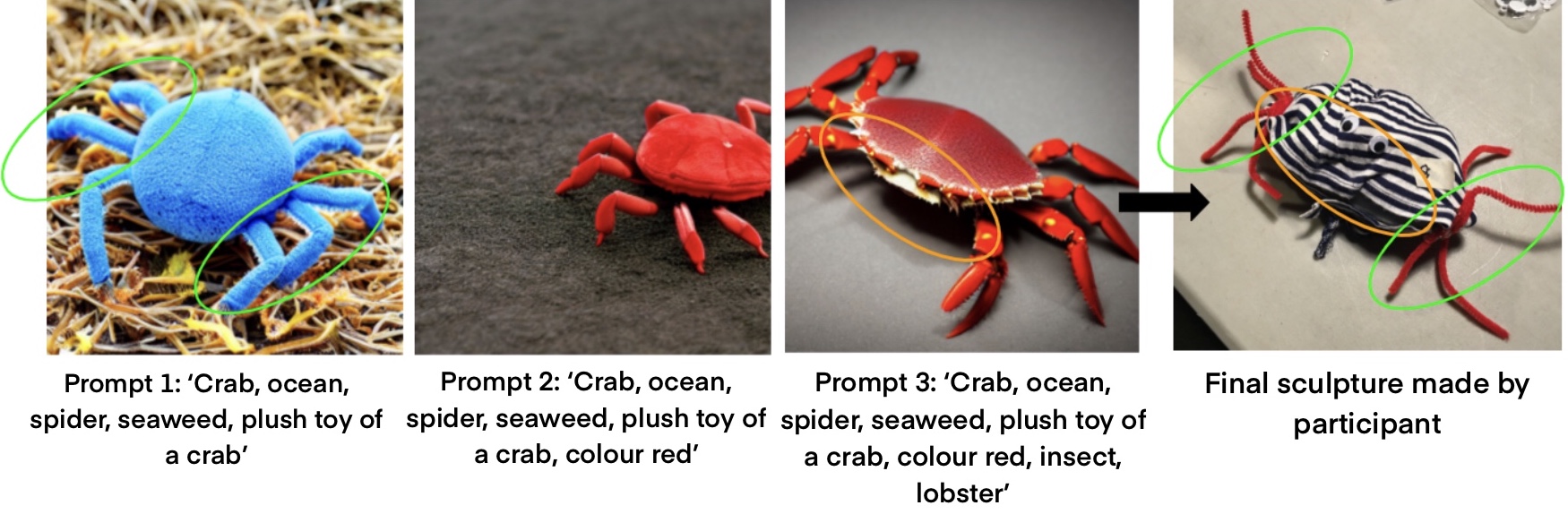}
  \label{fig:crab}
  \end{figure}

\begin{figure}[h]
  \caption{Visual elements relate to sculpture design of a garden.}
  \centering
  \includegraphics[width=85mm,scale=0.4]{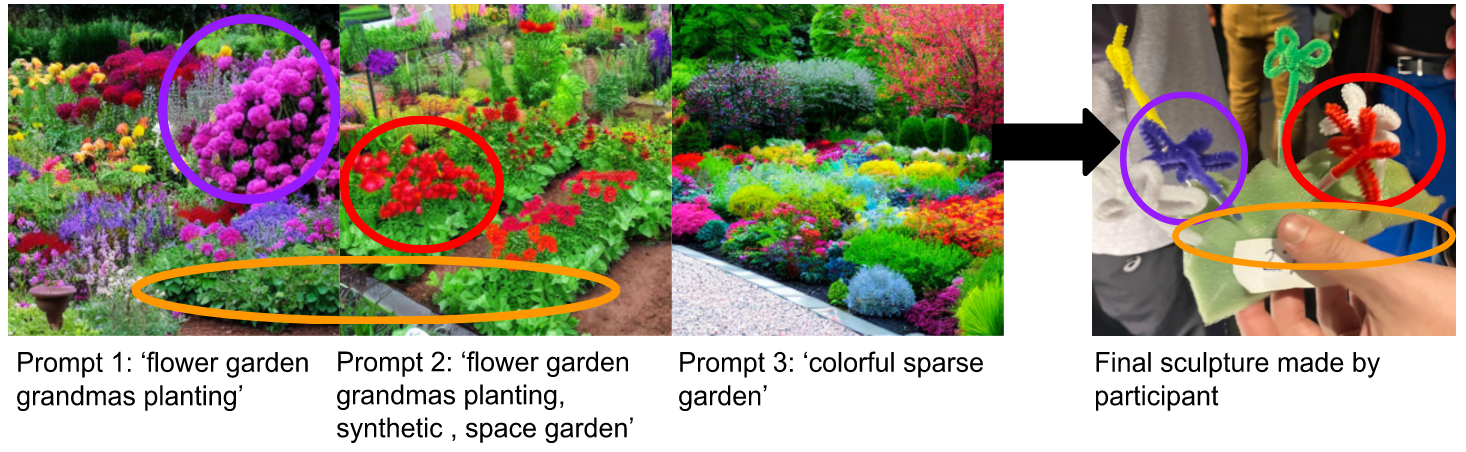}
  \label{fig:garden}
  \end{figure}

\begin{figure}[h]
  \caption{Visual elements inform sculpture design of a robot.}
  \centering
  \includegraphics[width=85mm,scale=0.4]{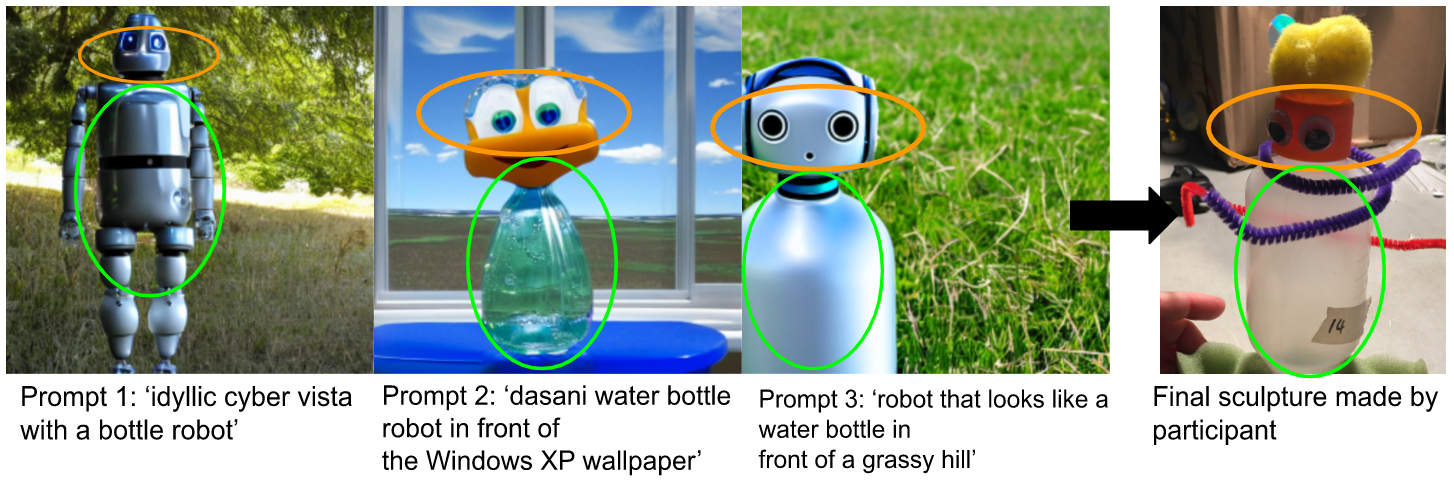}
  \label{fig:robot}
  \end{figure}

Of the 30 participants, 27 produced at least two prompt and image pairs. Of those 27, 24 produced all three images (the remaining three produced just one image).
23/30 participants self-reported that their sculptures were informed by the images they saw, and 28/30 participants reported that they would use text-to-image models again for a creative task. Figures~\ref{fig:building}, ~\ref{fig:crab}, ~\ref{fig:garden}, and ~\ref{fig:robot} show examples of participants' three prompt and images pairs and the created sculptures that have strong visual links to the final design highlighted.

Figure ~\ref{fig:building} shows an example of strong image influence on sculpture design for someone who did not have a sculpture idea going into the process. The rightmost image is a photo of the sculpture they built, which is striking in its visual similarity to the third generated image in particular. Some congruences, including color of materials which were not changeable, are highlighted in colored boxes on the image. 

Figure ~\ref{fig:crab} shows the journey of someone who knew from the onset they wanted to make a sculpture of a crab. The three images are outputs from the prompts: ``crab, ocean, spider, seaweed, plush toy of a crab" \textrightarrow ``crab, ocean, spider, seaweed,  plush toy of a crab, colour red" \textrightarrow ``crab, ocean, spider, seaweed, plush toy of a crab, colour red, insect, lobster." Despite already knowing what they wanted to make, the participant said the images gave reminders of a crab's features so they could create it in real life. The participant tweaked their second prompt to include ``colour red" because their physical materials were also red. Common entities like crabs are easy to generate thanks to their high visual determinacy, but some users with more abstract ideas than the example discussed above reported frustration visualizing what they had in mind, even if they knew what they wanted to see in the images. 

Of the participants who reported that the images were not informative to their sculpture (n = 7), three explained that this was because images did not sufficiently relate to the materials given and they could not build what they saw. Some of these participants expressed frustration in formulating a text prompt that would produce what they wanted to see in an image. Only 36.2\% of images contained elements that the participants expected to see, indicating many visual elements were unexpected. 40\% of participants (12/30) directly referenced their chosen materials, like ``box" or ``sponge." \footnote{We do not find statistically significant differences in the distribution of average cosine distances of user prompts and whether or not the prompt contained mentions of physical materials (t = 1.01, p = 0.32), material qualities (t = -1.25, p = 0.22), or colors (t = 0.12 , p = 0.91).} The frustration some users had translating between materials and expected images suggests a need to better support users in translating between physical materials and prompt wording when ideating for physical artefacts.

In post-interviews, the reasons participants reported that images were informative were diverse. Some found that the lateral concepts introduced by the images gave them new conceptual ideas, like the addition of a river in Figure  ~\ref{fig:building} perhaps indicating they were in an early design stage where exploration was particularly useful. Others honed existing ideas by getting implementation inspiration from the images, like in Figure ~\ref{fig:crab} perhaps indicating a later design stage focused on execution. These findings confirm those in \citet{epstein2022happy}, which showed that AI-generated images are helpful in visualizing ideas for two main reasons: they give lateral insight as well as implementation ideas.

Two participants who did not find the images helpful built their sculpture immediately after their first image, and explained that the strength of their pre-visualization ideas meant that they believed no images would further inform the sculptures they would make. This might suggest that seeing images is less helpful for users with an already concrete vision, but we do not find strong evidence to support that claim. Of the individuals with a design idea before starting the activity, 3/8 found seeing the images helpful and 5 did not. Of the individuals with no design idea before, 15/23 found seeing the images helpful and 8 did not, a 44\% difference. This provides suggestive evidence (t = -1.365, p = 0.183) that participants who did not have an idea may find the generated images more helpful. 

\subsection{Distinct prompting styles emerge} 
Participants' wording choices for prompts ranged in their degree of conceptual exploration. We observed and named some patterns we saw in users' prompting choices. The most minimal amount of conceptual exploration that we observed was a pattern we call the ``refiner" style. In these instances, the participant started with a prompt and made minor edits to it. It was the most common pattern we saw (n = 15). For example: \textbf{Prompt 1:} `geometric ominous dystopian creature metallic'. \textbf{Prompt 2:} `geometric ominous dystopian creature \emph{neon heart beating}'. \textbf{Prompt 3:} `geometric ominous creature neon heart beating \emph{rainbow}'. The participant described here kept changing the final terms to try to see the image they wanted.

The next pattern observed was the ``rephraser" prompting style (n = 5). Here, conceptual subject matter remained the same, but changes to the wording or order of the prompt changed substantially. Example: \textbf{Prompt 1:} `idyllic cyber vista with a bottle robot'. \textbf{Prompt 2:} `dasani water bottle robot in front of the Windows XP wallpaper'. \textbf{Prompt 3:} `robot that looks like a water bottle in front of a grassy hill'. Each prompt contains a reference to a robot, but the way they refer to the same background of a grassy hill changes.

A few participants did even more conceptual exploration. The ``explorer" prompting style (n = 2) we describe consisted of three conceptually unrelated prompts. Example: \textbf{Prompt 1:} `my printed map required too much plastic'. \textbf{Prompt 2:} `I'm exhausted but I'm still having fun'. \textbf{Prompt 3:} `my hovercraft is full of eels'. The small size of this category shows that most people had some idea of what they were exploring.

These described styles form an exploration gradient where ``explorers" have the most semantic distance traveled, ``rephrasers" have a main idea but still explore around it, and ``refiners" focus on exploitation of a single string of words.

We observed that a remaining 7 participants did not fit neatly in to these qualitatively defined prompting categories, often using a mix of styles we have described. For example, the user who built the sculpture in Figure \ref{fig:building} prompted as follows: \textbf{Prompt 1:} ``Hamburg is the most beautiful city in  the world", \textbf{Prompt 2:} ``sun and shiny water, colourful buildings made of trash and ropes", \textbf{Prompt 3:} ``green red and white buildings in front of a shiny river with a white fence on the side". The first prompt appears to be exploratory, whereas the second two describe a more concrete scene in two similar but distinct ways, reminiscent of behavior we see in those using a ``rephraser" style. The final three participants to account for only gave a single prompt, so their prompting style does not form a style that can be attributed over time.

Because not all participants fit neatly into the prompting styles we qualitatively described, we use a computational measure of semantic distance to objectively characterize all prompting journeys in the rest of our analysis. We calculate the average conceptual distance a participant traveled through their prompting journey by taking the cosine distance between the text embeddings of first and second prompts, then the second and third prompts, then averaging the two distances. For participants with two prompts, we treat the single distance between the two prompts as the "average" distance. For participants with a single prompt, where no distance between prompts was traveled, we defined the distance as 0. 

 \begin{figure}[h]
  \caption{Histogram of participants' conceptual exploration across their prompting session, measured by average cosine distance. Shown above histogram: mean cosine distance within a prompting style and 95\% confidence intervals. The prompting styles appear to occur at different average cosine distances. 
  }
  \centering
  \includegraphics[width=85mm,scale=0.10]{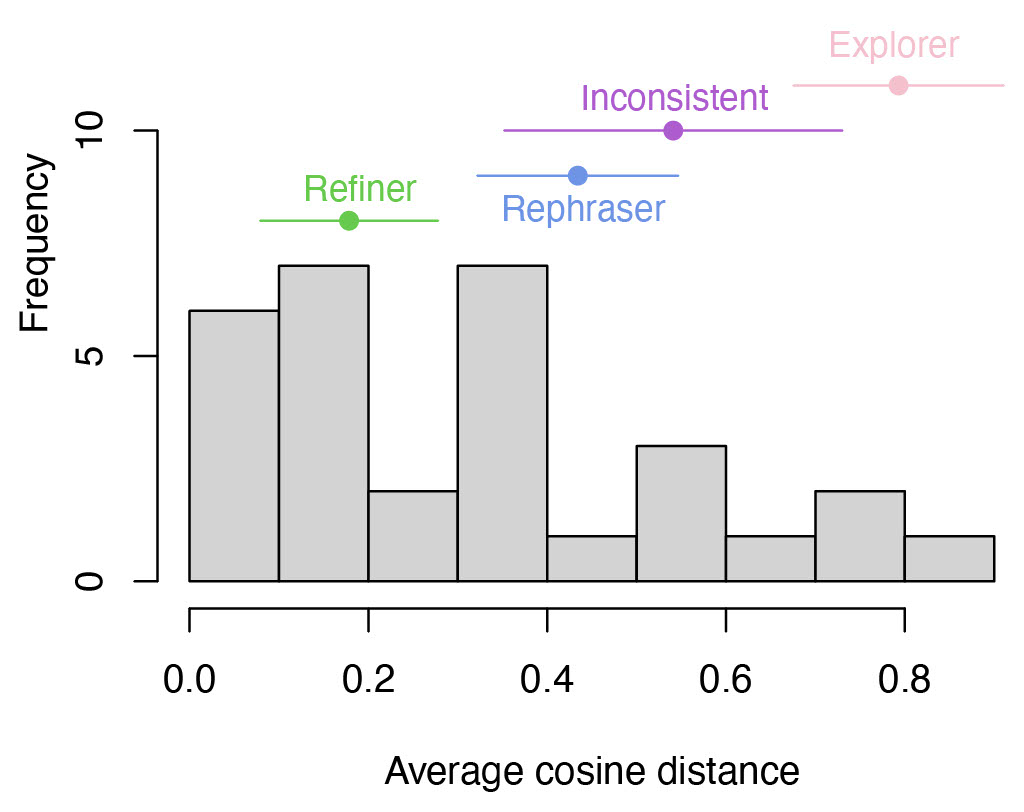}
  \label{fig:hist}
  \end{figure}

Figure \ref{fig:hist} shows a histogram of average cosine distances across the prompting journeys of our 27 participants who gave at least 2 prompts, along with the average and standard deviation of the subgroup we qualitatively grouped into ``refiner,'' ``rephraser,'' ``inconsistent,'' and ``explorer.'' The participants we describe as ``refiners" have the lowest average cosine distance of 0.178, while the participants we described as ``explorers" had the highest average cosine distance of 0.793. The participants we describe as ``rephraser" and ``inconsistent" fall in the middle range.

\begin{figure}[h]
  \caption{Prompts and image pairs for the different prompting styles illustrated in Figure 3. From top to bottom: `refiner', `rephraser' and `explorer'}
  \centering
  \includegraphics[width=85mm,scale=0.2]{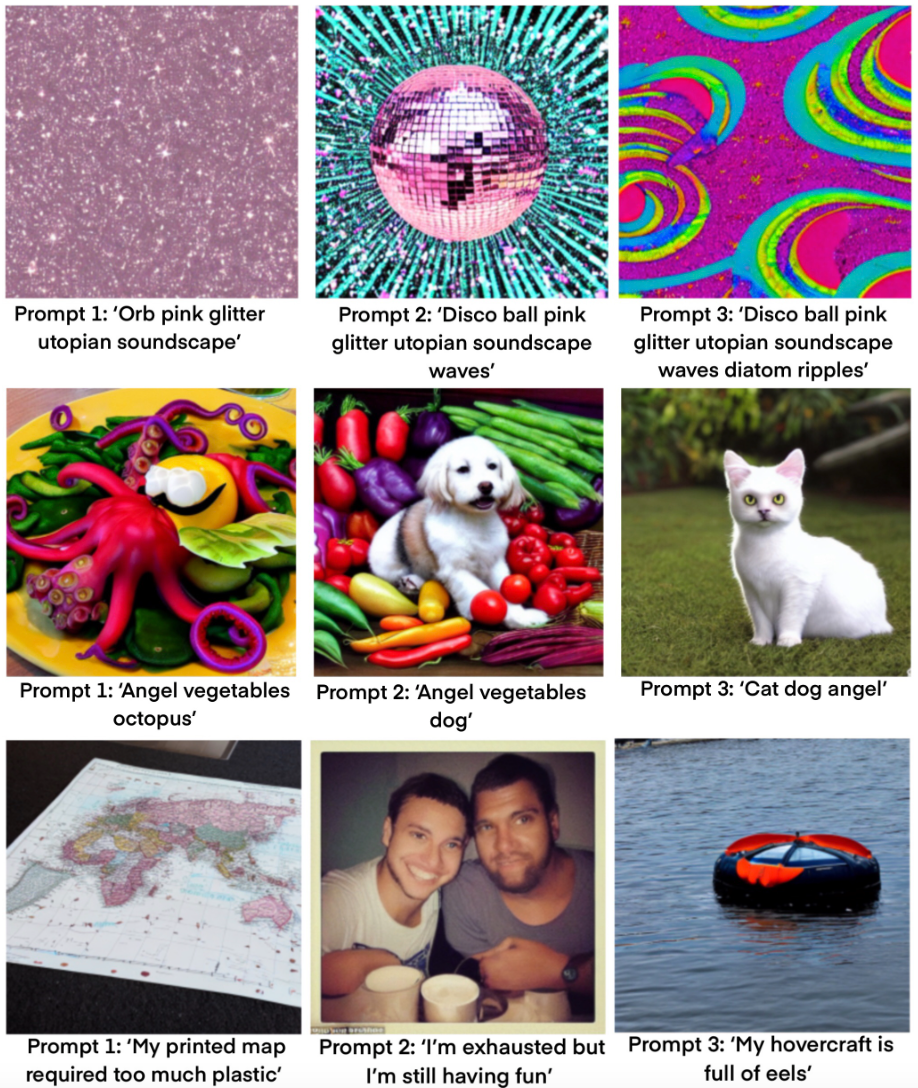}
  \label{fig:allUsers}
  \end{figure}
\begin{figure}[h]
  \caption{TSNE plot of embedded representations of three user prompting sessions where users range in degree of conceptual exploration while prompting}
  \centering
  \includegraphics[width=85mm,scale=0.14]{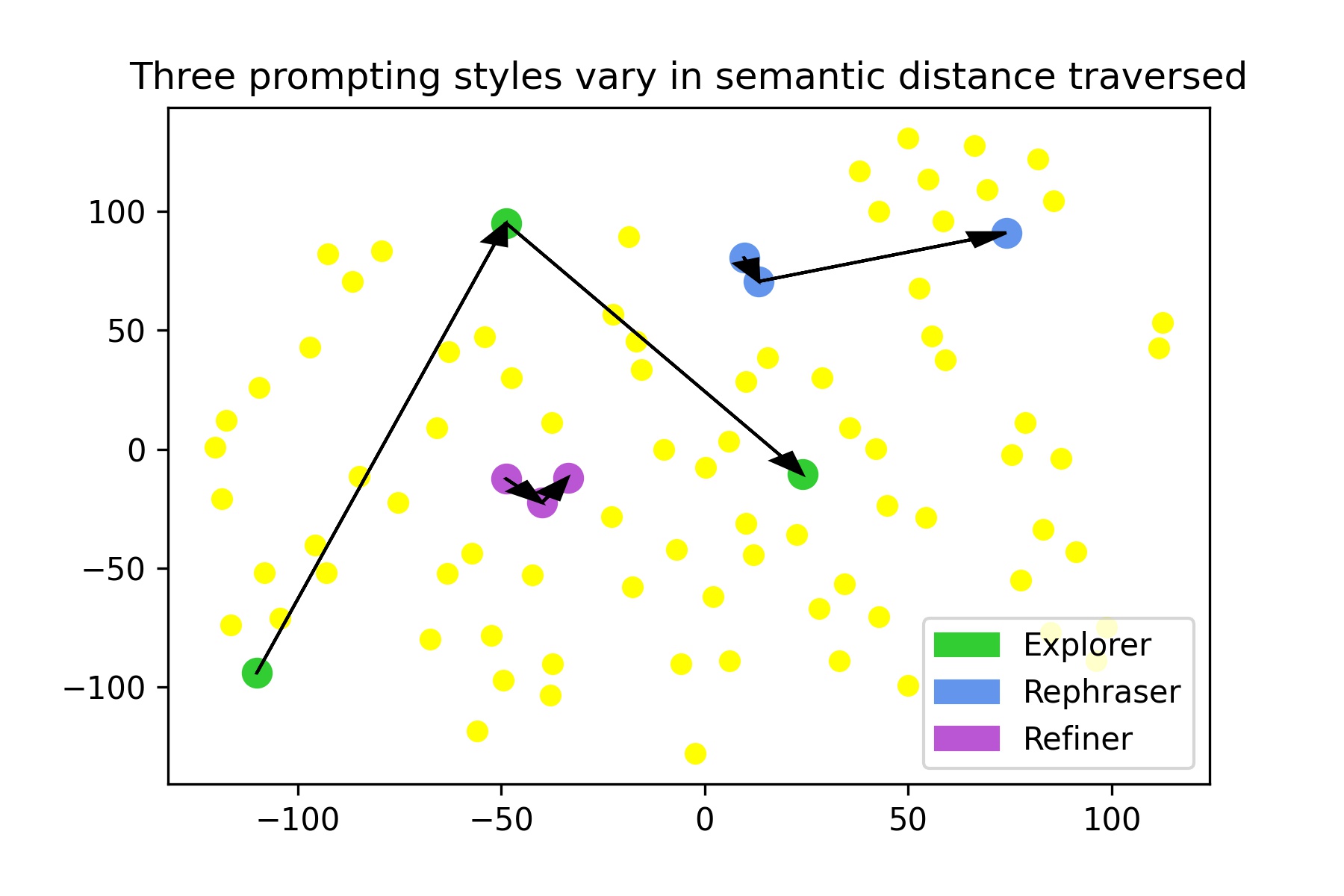}
  \label{fig:distance}
  \end{figure}

We visualize three participant journeys, one from each of the main styles we have described, through semantic space in Figure \ref{fig:distance}. We use TSNE to reduce dimensionality of the shared embedding space to two dimensions, and draw arrows from points representing a participant's first to second to third prompts. The path between green dots in Figure \ref{fig:distance} (representing a single user's prompting journey) shows considerable conceptual exploration using an ``explorer" prompting style. The participant journey represented by blue dots shows a less exploratory ``rephraser" style, and the user represented by purple dots shows the ``refiner" prompting style. The prompts and corresponding images generated for these three participants can be seen in Figure \ref{fig:allUsers}. 

\subsection{Prompting Style differs by Design Stage} 

We next investigate the relationship between the design stage a user indicated, based on whether they described in our post-interview having started the task with a sculpture idea or not, and the average semantic distance they traveled. 
Because we gave participants the design task of creating a sculpture, participants skipped what \citet{angel} calls the ``Q\&A" stage of defining the creative task. By constraining the category of materials and type of creative output (sculpture), we further define the design problem's ``artefact type" and ``media type," per the \citet{pip} conceptualization of the design process. 
Those who came into the visualization stage of the exercise with no idea what to make started in the ``wandering stage," wherein creatives explore possible strategies and incubate ideas. Some who started visualization with a sculpture idea skipped this stage and were already in the ``hands-on stage" of developing solutions, or were already moving to the ``camera-ready stage," which focuses on selecting and implementing ideas, to use \citeauthor{angel}'s taxonomy. Using these stages of the creative process, we compare the average semantic distance traveled for the group of participants who had sculpture ideas before visualizing with generative AI to those who did not.

We found that the amount of conceptual exploration a participant did was lower for participants who said they had a sculpture idea at the visualization stage than those who did not (t = -2.94, p = 0.006). We also found that participants who stopped early, generating fewer than 3 images, employed less conceptual exploration in their images than those who did not (t = 4.31, p \textless 0.001). All participants who stopped after one (n = 3) or two images (n = 4) had an idea of what to make before starting their visualization. 

\section{Future Work and Limitations}

Interesting preliminary findings emerged from this pilot study, but these findings could be investigated more precisely with changes to the study design that are informed by this first experiment.
A similar experiment that allows participants to choose their materials after visualization, instead of before, could investigate whether there is a relationship between prompting styles and the objects people select to build with. Similarly, one could imagine comparing two groups, one that used Stable diffusion for ideation, and one without access to the tool. 

We could also ask participants upfront whether they have a sculpture idea or not before their prompting journey starts, as in this pilot study the design stages recorded by facilitators were determined by participants' post-interviews and this data was therefore emergent. Participants' comments reflecting on their design stages in their post-interview may have been affected by seeing the generated images already. 
We also gave participants a maximum of three tries with Stable Diffusion in this experiment for the sake of time, but more complex prompting behavior may emerge for users across more interactions with the technology. Anecdotally, we also saw that some users were distracted by the technology itself, so future work could directly observe the effect of user familiarity with the tool on the user's prompting journey, and we are unsure how user familiarity with the tool may have affected the prompting journeys we describe.

The embedding representations we chose for prompts could also use the CLIP or BERT tokenizers that were used in Stable Diffusion for greater conceptual similarity to the tokenization process of text to image models, and a sensitivity analysis of individual tokens on these embeddings could be done to deepen conclusions.

\section{Conclusion}

Most participants found seeing the images generated by a text-to-image model informative to their final design of a sculpture. The average semantic distance a participant traveled in their prompting journey during the visualization activity was lower if they had a sculpture idea to start the activity than if they did not. This shows that prompting decisions relate to a participant's design stage. Participants who started visualization with sculpture ideas already in mind used image generation as an opportunity to ``exploit" or refine ideas, traveling less average semantic distance than those who were unsure what to build and used the images to explore. To better support creators,  text-to-image tools could identify a user's semantic distance traveled over a prompting session to suggest hints that are useful to their current design stage.

\section*{Acknowledgements}
We thank Pip Mothersill and Janet Rafner for helpful comments and feedback. We thank the 99F crew, participants of an earlier iteration of this workshop, and Kim Hew-Low for help shaping the workshop.



The vast majority of participants in our study were interested in using a generative creative tool like this, and they had many ideas for ways they would do so---- from designing a new kind of battery to generating artwork for their bedrooms. With thoughtful prompting guidance,  we hope that a wide of variety of creators can use text-to-image models to inspire visionary ideas and bring those ideas from our screens into reality.


 
%

\bibliography{aaai23.bib}



.


  

\end{document}